\documentclass[letterpaper]{article} 
\usepackage{aaai23}  
\usepackage{times}  
\usepackage{helvet}  
\usepackage{courier}  
\usepackage[hyphens]{url}  
\usepackage{graphicx} 
\usepackage{natbib}  
\usepackage{caption} 
\usepackage{algorithm}
\usepackage{algorithmic}

\usepackage{newfloat}
\usepackage{listings}
\DeclareCaptionStyle{ruled}{labelfont=normalfont,labelsep=colon,strut=off} 
\floatstyle{ruled}
\newfloat{listing}{tb}{lst}{}
\floatname{listing}{Listing}
\usepackage{subfiles} 
\graphicspath{{figures/}{../figures/}}
\usepackage{subfig}
\usepackage{multirow}

\title{Eloss in the way: A Sensitive Input Quality Metrics for Intelligent Driving}
\author{
    Haobo Yang, Shiyan Zhang, Zhuoyi Yang, Xinyu Zhang\thanks{Corresponding author} \\
    \texttt{s1911593@ed.ac.uk, zshiyan@bupt.edu.cn,} \\
    \texttt{zhuoyiyang03241811@tju.edu.cn, xyzhang@tsinghua.edu.cn,} \\
}
\affiliations{

    The State Key Laboratory of Automotive Safety and Energy, \\
    and the School of Vehicle and Mobility\\
    Tsinghua University\\
}

\usepackage{bibentry}

\begin{document}

\maketitle

\begin{abstract}

    With the increasing complexity of the traffic environment, the importance of safety perception in intelligent driving is growing. Conventional methods in the robust perception of intelligent driving focus on training models with anomalous data, letting the deep neural network decide how to tackle anomalies. However, these models cannot adapt smoothly to the diverse and complex real-world environment. This paper proposes a new type of metric known as Eloss and offers a novel training strategy to empower perception models from the aspect of anomaly detection. Eloss is designed based on an explanation of the perception model's information compression layers. Specifically, taking inspiration from the design of a communication system, the information transmission process of an information compression network has two expectations: the amount of information changes steadily, and the information entropy continues to decrease. Then Eloss can be obtained according to the above expectations, guiding the update of related network parameters and producing a sensitive metric to identify anomalies while maintaining the model performance. Our experiments demonstrate that Eloss can deviate from the standard value by a factor over 100 with anomalous data and produce distinctive values for similar but different types of anomalies, showing the effectiveness of the proposed method. Our code is available at: (code available after paper accepted).

\end{abstract}

\section{Introduction}

    \begin{figure}[t]
      \centering
      \includegraphics[width=0.95\columnwidth]{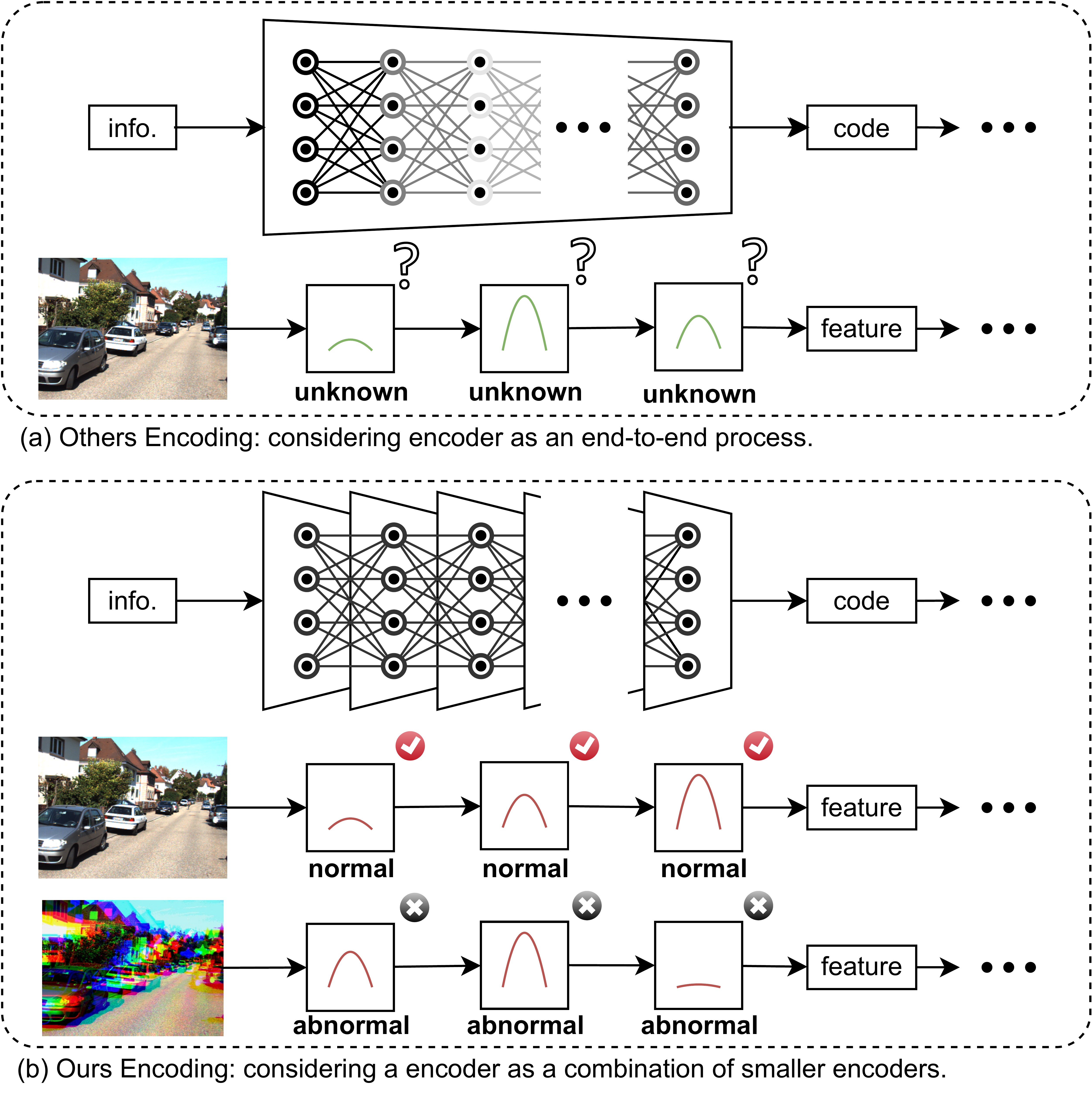}
      \caption{Pending update}
      \label{distribution-figure}
    \end{figure}
    
    Intelligent driving is an inevitable trend in the future development of urban transportation \cite{freudendal2019smart}, of which 3D object detection is one of the essential tasks to achieve intelligent driving \cite{zhi2017lightnet}. In recent years, with the rapid iteration of computing devices and the emergence of high-quality annotation datasets, data-driven deep learning methods have driven the rapid development of 3D object detection tasks \cite{wu2021multi}. However, compared with the target detection tasks applied to other fields, because of driver and road safety, the 3D object detection tasks applied to intelligent driving have higher requirements for accuracy and speed. In addition, compared with other application scenarios, the sensors in the intelligent driving vehicle can not guarantee the collection of stable and high-quality data due to uncontrollable conditions, for example, the weather. As a result, some data may appear anomalous. If a model can not judge these anomalous data and make special treatment, it would lead to errors in the decision-making of the intelligent driving system and even severe traffic accidents. 
    
    In the face of the higher requirements of intelligent driving for 3D object detection, many advanced algorithms have given their solutions to achieve higher accuracy, and faster speed. However, much of this work is based on high-quality data modeling, focusing on achieving higher performance on clean datasets. Furthermore, even if there is a small range of low-quality data in the dataset, the model can learn several anomalous patterns through large-scale training, thus achieving high accuracy. However, the data collected in the real world is diverse, and the patterns learned in the dataset are not enough to cope with the complex driving environment. Therefore, even if high accuracy is achieved on various data sets, the perception model will most likely produce wrong judgments and make wrong decisions, resulting in irreparable consequences.

    Anomaly detection has solutions in many fields, such as CCD flatness detection and textile defect detection. However, much of this work is extensively based on data, training a “black box“ model, which is unsuitable for intelligent driving. Moreover, as mentioned earlier, models can only learn a limited number of anomalous patterns through training data, and limited patterns are not enough to cope with the complex and diverse road environment. Therefore, to solve the problem of anomaly, we should look for the difference between normal and anomalous from the data itself. The ultimate purpose of a network is to extract helpful sensor information regarding the target task; for those anomalous data, the network cannot extract adequate information from it or even over-interpret these data. From this starting point, we can detect the anomalous data in real-time according to the amount of information extracted by the model or by each model layer, see Figure~\ref{distribution-figure}.
    
    In this paper, from the perspective of treating the perceptual system as a communication model \cite{zou2022novel}, we build a theoretical model that can describe the underlying information transmission process of the neural network for intelligent driving perception tasks. Our contribution mainly includes the following aspects. Firstly, based on the source coding theory in the communication system \cite{jones_2000_information}, the expected value of information change at each layer of the information compression part of a model has been constructed: the information change at each layer of the information compression network is stable. Second, a probabilistic model is established for the data in the neural network by introducing a continuous random variable $X$, to estimate the change in information entropy. Third, according to the expected value of information change, the plug-and-play Eloss function module is established, which can be used as a loss function to guide the process of neural network parameters’ update and give networks the ability to detect the anomaly.

\section{Related Works}

    \subsection{Uncertainty quantiﬁﬁcation}
        In real-world scenarios, autonomous driving perception systems face challenges such as occlusion and noise in sensor observations [1], which are often limited in the observability of the environment. Therefore, the quantification of perceived uncertainty in various environments has attracted attention.
    
        Deep learning uncertainty has two types, Aleatoric Uncertainty and Epistemic uncertainty [2]. Aleatoric uncertainty is due to an incomplete understanding of the environment, such as partial observability and measured noise, which cannot be reduced by obtaining more or even unlimited data but can be reduced by explicit modeling. Aleatoric uncertainty is usually learned using the heteroscedasticity loss function [3]. Epistemic uncertainty comes from a deficient dataset and some methods unknown to our model, which can be eliminated with enough training data. Two popular methods usually estimate Epistemic uncertainty: Monte Carlo (MC)-dropout [4] and ensembles [5]. 
        
        In previous studies, uncertainty quantification lacked the truth value of uncertainty estimates [6] and a unified quantitative evaluation index [7]. More specifically, uncertainty is defined differently in different machine learning tasks, such as classification, segmentation, and regression. In the paper, the proposed method can produce a ground truth of 0 for regular data input, and our Eloss is a unified method for all kinds of tasks or networks where a repetitive structure is presented. We discuss this in the next section.


    \subsection{ Shannon's source coding in the communication model}
    
        
        We can use communication models to build neural networks, using prior knowledge from information theory to explain and guide neural network optimization. 
        
        Information theory, mainly constructed by Shannon \cite{jones_2000_information}, uses the concept of entropy to study information processes based on information quantification. Information entropy reveals the limitations of signal processing and communication operations by quantifying the uncertainty in a signal. Due to the convenient quantization nature of communication networks, many have introduced knowledge of information theory in communication into deep neural networks.
        
        A channel model built by neurons or networks was suggested by MacKay et al \cite{mackay_2003_information}. Sharma et al. introduced fiducial coding in variational self-encoders \cite{sharma_2021_dagsurv}. Using Shannon's first theorem, the average length of the encoding is compared to the magnitude of entropy to ensure that the variational distribution is a distortion-free coding process. 
        
        In 2015, Tishby and Zaslavsky used IB \cite{tishby_2000_the} to further reveal the deep learning model mechanism. Their experiments using a multilayer perceptron showed that the network tended to capture relevant information first and then combine them. Inspired by the work of DAGSurv\cite{sharma_2021_dagsurv}, we modelled the pre-fusion process of feature extraction for a single modality as source coding and achieved compression of the amount of information and improved model efficiency by jointly training this part with the fusion and detection networks that follow, selectively capturing effective features and reducing unincrease features, while introducing entropy from information theory to quantify the amount of information in the output of each layer of source coding, by limiting the entropy changes in entropy to reduce the direction of possible optimization of the network and accelerate model optimization.
        
    \subsection{Optimizer of Neural Network}
    
        The process of optimizer optimization in machine learning is to look for neural network parameters that significantly reduce the loss function, which typically includes performance metrics evaluated on the entire training set and additional regularization terms. In order to make the model output approximate or reach the optimal value, we need to use various optimization strategies and algorithms to update and calculate the network parameters that affect model training and model output. Currently, the main solutions to this problem are roughly divided into three categories: gradient descent method, momentum optimization method, and adaptive learning method.

        In the gradient descent method, the small and medium batch gradient method combines the advantages of BGD\cite{hinton2012neural} and SGD\cite{bottou2012stochastic}, and selects less than the total number of training samples of small batch samples in sample selection, which not only ensures the speed of training, but also ensures the accuracy of the final convergence\cite{ruder2016overview}.
        
        Momentum optimization methods introduce momentum ideas in physics, accelerate gradient descent, and common algorithms are Momentum and NAG. Using the method of momentum optimization, it is possible to make the direction of the gradient in the unchanged dimension, the parameter update becomes faster, and when the gradient changes, the update parameter becomes slower, so that the convergence can be accelerated and the turbulence can be reduced\cite{dozat2016incorporating}. Among them, NAG is an improvement of momentum, which reserves the direction of the previous update when the parameters are updated, and uses the current gradient to fine-tune the final update direction, while introducing a correction at the time of gradient update. This method is very good at accelerating convergence and suppressing oscillations.
    
        In deep learning, the learning rate, as a very important hyperparameter, is generally difficult to determine and usually requires a certain number of trainings to find the optimal learning rate\cite{smith2017cyclical}. The adaptive learning rate optimization algorithm can adaptively adjust the learning rate size according to some strategies, thus improving the training speed. Currently, adaptive learning rate optimization algorithms mainly include: AdaGrad algorithm, RMSProp algorithm, Adam algorithm, and AdaDelta algorithm\cite{le2011optimization,zaheer2019study}. The momentum in Adam is directly incorporated into the estimation of first-order gradient moments. Likewise, Adam includes bias corrections to correct the first- and second-order moment estimates that are initialized from the origin. This makes the parameters relatively smooth and suitable for most nonconvex optimization problems, as well as large data sets and high-dimensional spaces.

\section{Theory}

    To efficiently fuse the information of each modality and remove the information that is not helpful to the subsequent task, the information must be compressed so that the features entering the subsequent network can contribute to the final task. This process of information compression in the communication model can be expressed as a distortion-limited encoding. To improve information transmission efficiency, source symbols that appear less frequently in the original source need to be removed during the compression process using distortion-limited encoding.
    
    Because these source symbols appear less frequently, even if they are lost during the encoding process, a high data recovery rate can still be achieved at the other end of the channel and transmission efficiency is greatly improved. Similarly, in an information compressing network, the removed information usually does not contribute to subsequent tasks, and removing this information can greatly improve the efficiency of following networks\cite{jones_2000_information}. 
    
    In communication models and information theory, entropy is often used as a measure of the amount of information. The lower the entropy, the larger the amount of information. As a result, we introduced an entropy calculation method to describe the change in information during distortion-limited encoding, which will be explained in detail later. At the same time, to ensure the smoothness of information compression, we keep the entropy change constant as the information passes through the layers of the neural network, preventing sudden distortion of the information.

    \subsection{Entropy Expectation of Neural Network Layers}
        \begin{figure*}[t]
            \centering
            \includegraphics[clip=true, width=0.99\textwidth]{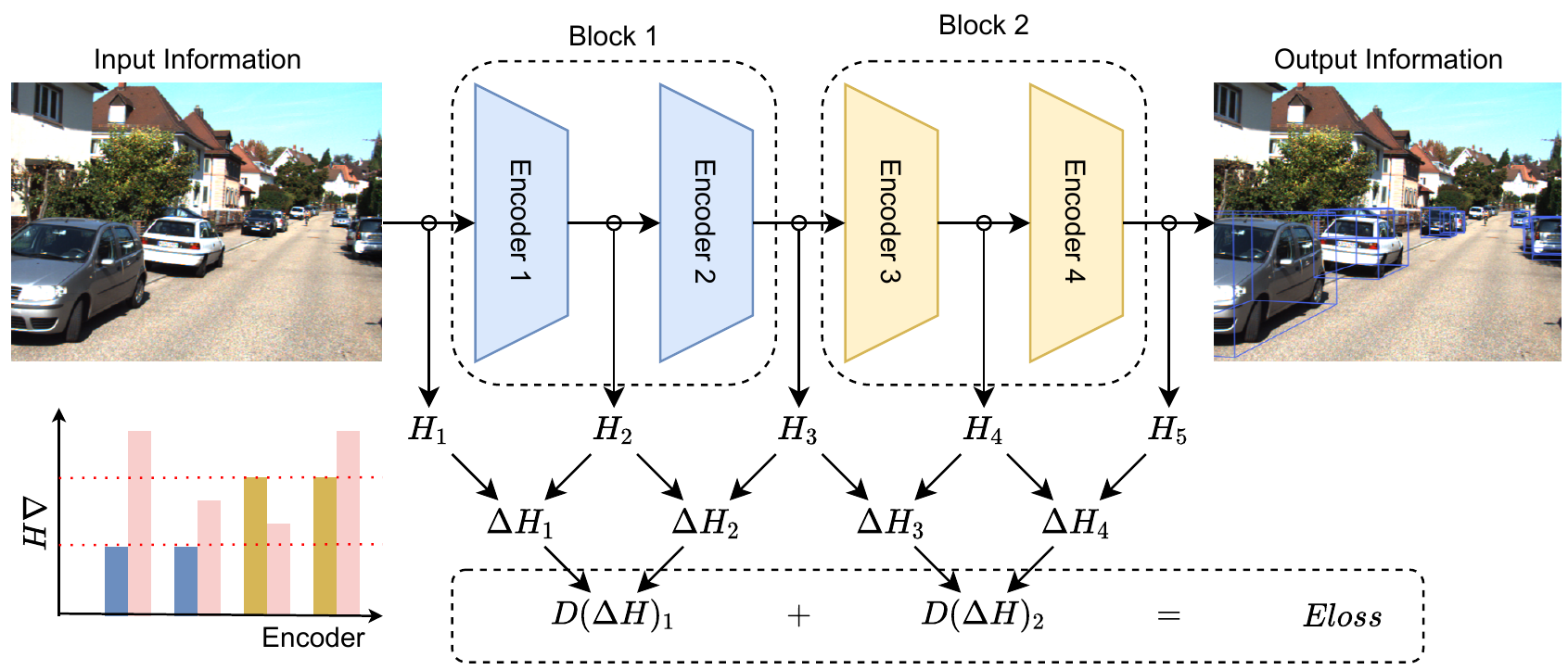}
            \caption{An detailed Illustration of our proposed Eloss method.}
            \label{fig1}
        \end{figure*}
        
        The neural network training process can be seen as constantly searching for the mapping relationship between input and expected output to achieve the effect of learning \cite{liu2017survey}. It can be considered that before the neural network is trained, the mapping relationship learned is weak, and when the neural network is trained with data, the mapping relationship learned is continuously strengthened, and the feature extraction ability of each layer of the network, that is, the information compression ability, is continuously improved. However, for such a black-box model\cite{buhrmester2021analysis}, there is no good explanation of the information compression process, and it is even more impossible to effectively guide the parameter optimization process of the information compression model.
        
        To explain the underlying mechanism of the information compression network, with reference to the communication model, feature compression can be regarded as a source coding process. To ensure that the network can extract data in a more complete way that is useful for subsequent tasks, the source encoding should be a distortion-limited encoding, as discussed above. 
        
        Now, we introduce the information entropy concept to reflect the amount of information output by the network. In a communication system with constant bandwidth, if the entropy of the information of transmitted data is steadily reduced, the efficiency of the information transmission is gradually improving \cite{zou2022novel}.

        In distortion-limited encoders, information entropy decreases with increasing degree of encoding, and the same applies to feature compression networks. Because the feature compression network often has a continuous repetitive network structure, such as the repeated linear layer in the SECOND network\cite{yan2018second}, and each duplicate network layer should have a similar feature compression ability, it can be considered that the bandwidth of data transmission in the network is constant. In summary, the expectation of the feature compression network for the change in the amount of information layer-by-layer is that the information entropy of the output results of each repeated layer of the network is steadily decreasing.
        
        According to the expectation, we can optimize the network parameters in the direction of keeping a steady change in the layer-by-layer information entropy by constructing the entropy loss function. Therefore, we are able to go into the black-box model and find a way to further optimize the training processes.
    
    \subsection{Uncertainty quantification for abnormal info input}
    
        For a feature extraction network with a specific function, the raw input data forms a new feature map after passing through each feature extraction layer, and in this process, there will be a change in the amount of information, that is, a change in information entropy. However, since no new information is introduced during feature extraction, and a large amount of information unrelated to the target task is filtered out, only the information related to the target task is retained. Therefore, this process is usually expected to be a process of information entropy reduction.
    
        In the same feature extraction network, the feature extraction capability of each layer of the network is positively correlated with the scale of the parameters of that layer. Therefore, when the feature map passes through several successive feature extraction layers with the same parameter amount and similar structure, a similar amount of information reduction should be maintained after each feature extraction to ensure the smooth information compression  of the network. 
        
        However, for an abnormal input feature map, for example, some information has noise, the change in the amount of feature map information when passing through each layer of the feature extractor will become irregular rather than a smooth decrease, and the entropy value may even increase. Therefore, we can observe the entropy of the input data as it passes through each feature extractor to determine whether the input information is an anomaly.

    \subsection{Probabilistic Modeling for Information}
        The loss function, calculated based on the expectation of entropy, requires a method to estimate the entropy of each layer output in the information compression network. Moreover, estimating the entropy requires probabilistic modeling of the distribution of output data.
        
        In information compression, it is common to consider the convolutional network as a choice \cite{zou2022novel}. To model the convolutional neural network as a probabilistic model, we set it as follows. The feature channels $ \tilde{X}=\{ x_1, x_2,... ,x_i \}$ generated by the convolution kernel are considered as samples of multidimensional continuous random variable $X$, where $i$ is the number of channels, and the number of values in each channel is the dimension $d$ of $X$.
        
        Because the output of each layer of any network can be considered a continuous random variable $X$, and the output of any layer can be selected as a set of samples $x_i$ of $X$, probabilistic modeling can also be applied to other neural network structures in addition to the convolutional neural network, to calculate the entropy of the data distribution. Therefore, the proposed probabilistic modeling method for convolutional neural networks can be extended to other neural network structures.
    
    \subsection{Entropy calculation}
        Now, the problem of estimating the entropy of each layer of the neural network is transformed into estimating the entropy according to the probability distribution of the unknown continuous random variable $X$.
    
        There are many ways to solve this problem, the essence of which is to calculate the differential entropy of continuous random variables. Differential entropy, also known as continuous entropy, is a concept in information theory that derives from Shannon's attempt to extend his concept of Shannon entropy to a continuous probability distribution \cite{jones_2000_information}. Set the random variable $X$, whose domain of the probability density function $f$ is the set of $X$. This differential entropy $h(X)$ or $h(f)$ is defined as follows:
        
        \begin{equation}
            h(X)=-\int f(x)\log{f(x)}dx
        \end{equation}
        
        Since the probability distribution of the random variable is not known in advance in this problem, the probability density function is unknown, and only a limited number of sample values are available in that probability distribution. Here is the K-Near-Neighbor Entropy Estimation Method \cite{van1988generalized} that we can use to calculate its information entropy: 
        
        Continuous variables are discrete with sampling, and to use $n$ samples to approximate the entire sample space, each sample point is expanded into a d-dimensional hypersphere with the radius of the sphere as the distance between the sample point and the nearest sample point. When the variables are evenly distributed in the sample space, the probability of each sample point can be approximated to $1/n$.
        
        Since the distribution of random variables in the sample space is unknown, there may be large differences from the uniform distribution, and the distribution of random variables in the space is corrected by using the distribution of the sample in the space. The density and sparsity of the samples in the sample space directly affect the probability density near each sample point. The discrete probability of each sample point is estimated as:
        
        \begin{equation}
            p(x_i)=[(n-1)\cdot r_d(x_i)^d]\cdot V_d]^{-1}
        \end{equation}
        
        where $n$ is the number of samples, $r_d(x_i)$ is the Euclidean distance of d-dimensionality between the sample $x_i$ and its nearest sample point, and $V_d$ is the volume of the unit sphere in d-dimensional space.
        
        The estimate of the entropy of the random variable $X$ is:
        
        \begin{equation}
            H(X)=\frac{1}{n}\sum_{i=1}^n[-\log{p(x_i)}]+\gamma
        \end{equation}
        
        where $\gamma$ is the Euler-Maseroni constant, which is approximately equal to 0.5772.
        
        The K-Near-Neighbor Entropy Estimation Method expands the distance between each sample point and its nearest sample point to the distance to the k-th sample point closest to it, and the entropy estimate of the random variable $X$ becomes:
        
        \begin{equation}
            H(X,k)=-\psi(k)+\psi(n)+\log{V_d}+\frac{d}{n}\sum_{i=1}^n\log{r_{d,k}(x_i)}
        \end{equation}
        
        where $\psi$ is the Di-gamma function, $\psi(1)=-\gamma$, $\psi(n)\sim\log{(n-1)}$. $r_{d,k}(x_i)$ is the Euclidean distance of d-dimensionality between sample $x_i$ and its nearest kth sample point. It can be shown that $H(X)$ is equivalent to $H(X,k)$ when $k=1$. We use $H(X)$ as the entropy value of the output of each layer of the network and then obtain the entropy variable $\Delta H$ of each layer of the network, $\Delta H_n=H_{n+1}-H_n$, where $n$ is the index of the layer of the network.
    
    \subsection{Loss Functions for Information Compression Network}
        According to the information transmmision expactation for a information compression network, there are two loss functions that can be established: function $L_1$ focuses on the steady change of information and function $L_2$ focuses on the direction of information change. 
         
        The loss function $L_1$ is formulated with the variance of the change in entropy $\Delta H$, and the expected value of $L_1$, the variance, is 0.
        
        \begin{equation}
            L_1=\frac{\sum_{n=1}^N(\Delta H_n-\widehat{\Delta H})^2}{N}
        \end{equation}

        Where $N$, $n$ is the index of the layer of duplicate network layers and $\widehat{\Delta H}$ is the mean entropy change of all duplicate network layers. 
        
        According to the decreasing entropy expectation, the loss function $L_2$ is formormulated as follows:
        
        \begin{equation}
            L_2=-\sum_{n=1}^N\Delta H_n^2
        \end{equation}
        
        We call $L_1$ and $L_2$ combined as Eloss. 
        
        As a network is trained with Eloss, the influence of Eloss can be described as an amplifier, enhancing the interpertability of some parts of the network, where it can be described as a feature compression network. But this is also a limit of the Eloss method, as the network layers where $L_1$ and $L_2$ can influence are limited to those that are repeated several times and perform a specific task closely linked to the communication network. We evaluate this limit in the following sections. For this reason, Eloss is a complement to the loss function $L$ that corresponds exactly to the final goal.

\section{Experiments}

    \textbf{Dataset.} We conduct experiments on the KITTI dataset set and nuScenes dataset, which is jointly founded by the Karlsruhe Institute of Technology in Germany and the Toyota Institute of Technology of America \cite{geiger_2012_are,geiger2013vision}. In addition to that, this dataset is one of the most widely used computer vision algorithm evaluation datasets in the scenario of intelligent driving. The data set contains real-world image data collected from scenes such as urban areas, villages, and highways. The data in the KITTI dataset contains multimodal data, such as lidar point clouds, gps data, right-hand color camera data, and grayscale camera images. The KITTI dataset is divided into training sets and test sets, where the training set contains 7481 samples, and the test set contains 7518 samples. The Nuscenes(NuTonomy Scences) dataset is a multimodal dataset on autonomous vehicle driving. It is the first large-scale dataset to provide a full set of sensor data for an autonomous vehicle, including six cameras, one liDAR, five millimeter-wave radars, GPS and IMU. It contains seven times more object annotations than the KITTI dataset. These include 1.4 million camera images, 390,000 LiDAR scans, and manually annotated 1.4m 3D bounding boxes for 23 object categories. The Nuscenes dataset consists of 1000 scenes, each equivalent to 20S of video, containing a wide variety of scenarios. On each scene, there are 40 key frames, which is two keyframes per second, and the others are sweeps. The key frames are manually labeled, and each frame has some annotations in the form of bounding box. Not only size, fenced, category, visibility, and so on. Existing autonomous driving datasets lack a full set of multimodal data for building autonomous driving sensing systems, which Nuscenes compensates for.

    \textbf{Implementation Details.} The experiment was carried out on the Nvidia RTX 3090 device and the model was built using PyTorch based on the MMDetection3D \cite{mmdet3d2020} framework. PyTorch has a wide range of deep learning applications and provides a large number of Python interfaces, making it very easy for the framework to call and use Python's own function packages. The auto-differentiation feature included in PyTorch has made it a very popular dynamic graph framework. MMDetection3D is an open source toolbox for 3D object detection based on PyTorch, and the models used in this article are developed using PyTorch on the MMDetection3D framework.
    
    \subsubsection{Evalutate the Sensitivity of Eloss on abnormal inputs}
    \begin{table*}[!ht]
        \centering
        \resizebox{0.99\textwidth}{!}{
            \begin{tabular}{ll|l|ccc|ccc|ccc}
                \hline
                \multicolumn{2}{c|}{Setting} & \multirow{2}*{Value} & \multicolumn{3}{c|}{Confidence} & \multicolumn{3}{c|}{Eloss (metrics)} & \multicolumn{3}{c}{Eloss (metrics \& loss func.)} \\ 
                Model & Dataset & ~ & clean & noise1 & noise2 & clean & noise1 & noise2 & clean & noise1 & noise2 \\ \hline
                \multirow{2}*{VoxelNet} & \multirow{2}*{KITTI} & Mean & 0.495 & 0.248 & 0.248 & 0.015 & 0.008 & 0.009 & 1.584E-03 & 9.085E-03 & 8.697E-03 \\ 
                ~ & ~ & \%change & 0.0\% & -49.9\% & -49.9\% & 0.0\% & -48.5\% & -39.1\% & 0.0\% & \textbf{473.5\%} & \textbf{449.0\%} \\ \hline
                \multirow{2}*{PointPillars} & \multirow{2}*{KITTI} & Mean & 0.487 & 0.344 & 0.344 & 0.012 & 2.086 & 0.008 & 1.091E-01 & 3.836E+00 & - \\ 
                ~ & ~ & \%change & 0.0\% & -29.3\% & -29.3\% & 0.0\% & \textbf{17475.6\%} & -36.1\% & 0.0\% & \textbf{3416.0\%} & - \\ \hline
                \multirow{2}*{PointPillars} & \multirow{2}*{nuSenes} & Mean & 0.168 & 0.128 & 0.128 & 0.034 & 1.918 & 0.016 & 2.556E-04 & 1.478E-01 & 4.273E-03 \\
                ~ & ~ & \%change & 0.0\% & -23.7\% & -23.7\% & 0.0\% & \textbf{5494.7\%} & -54.5\% & 0.0\% & \textbf{57746.3\%} & \textbf{1571.9\%} \\ 
                \hline
            \end{tabular}
        }
        \caption{Comparison between Confidence, Eloss as metrics and Eloss as metrics and loss function. Observation with different noise setting is given.}
        \label{abnormal_detection_performance}
    \end{table*}
    
    The pointpillars model trained with and without eloss is used to process the KITTI data set and Nuscenes data set of the test data set respectively, and the voxelnet model trained with and without eloss is used to process the KITTI data set of the test data set respectively, and then the eloss that obtains the result is compared. Then calculate the confidence of processing the results of the data set in the model without eloss added. Compare the eloss of the experimental results with confidence of the experimental results, and further explore the difference between the eloss of the results obtained by the model trained with eloss and the results obtained by the data set compared with the model not trained with eloss, and explore which indicators can be used to better respond to abnormal data, and explain the difference in sensitivity to abnormal data. Our result is shown in Table~\ref{abnormal_detection_performance}.
    
    The confidence obtained by the model trained without adding eloss is compared with the eloss value after processing different test data sets. It is found that when processing data sets with different noises added under different models, the confidence after adding noise is lower than that when no noise is added. It can be identified that the data at this time is abnormal, but However, when the noise addition ratio is different, the confidence of the result changes very slightly, the sensitivity to data anomalies is low, and there is no discrimination, and the noise addition ratio cannot be judged. It is guessed that confidence has the problem of over-interpretation of adding noise. And the eloss value in this situation can also identify that the data is abnormal, which plays a role in the same effect, although the eloss value presents a chaotic state. However, it is particularly sensitive to some data with a specific proportion of noise, and can detect the addition of different sizes of noise. That is, when using the same model to obtain the same detection results, using eloss as an indicator can better judge that the data is abnormal, and can detect abnormal data with a degree of discrimination.
    
    The eloss obtained by the model trained without adding eloss after processing different test data sets is compared with the eloss value obtained by the model trained with adding eloss after processing different test data sets. It is found that the eloss value of the latter is more sensitive to data anomalies, and the fluctuation of the eloss value is more obvious, although the accuracy of the result is reduced. That is, when using the same model, the model trained with eloss will lose some accuracy in the result, but at the expense of a certain detection accuracy, the detection ability of abnormal data can be further improved.
    
    Comparing the eloss value with confidence in these two cases, it is found that although confidence can reflect that the data is abnormal, the change range is extremely small when different proportions of noise are added, and no further distinction can be made; However, the eloss value is used as an indicator to judge the degree of abnormality of the data, especially the eloss value obtained by the model trained by eloss. Although the detection accuracy has decreased, not only can the data abnormality be judged in the same way, but also the eloss fluctuation is larger and further improves the detection ability of abnormal data.

    \subsection{Comparison with Eloss on Training Process}
        \begin{figure*}[t]
          \centering
          \subfloat[Car $AP_{dist1.0}$]{
            \includegraphics[width=0.333\textwidth]{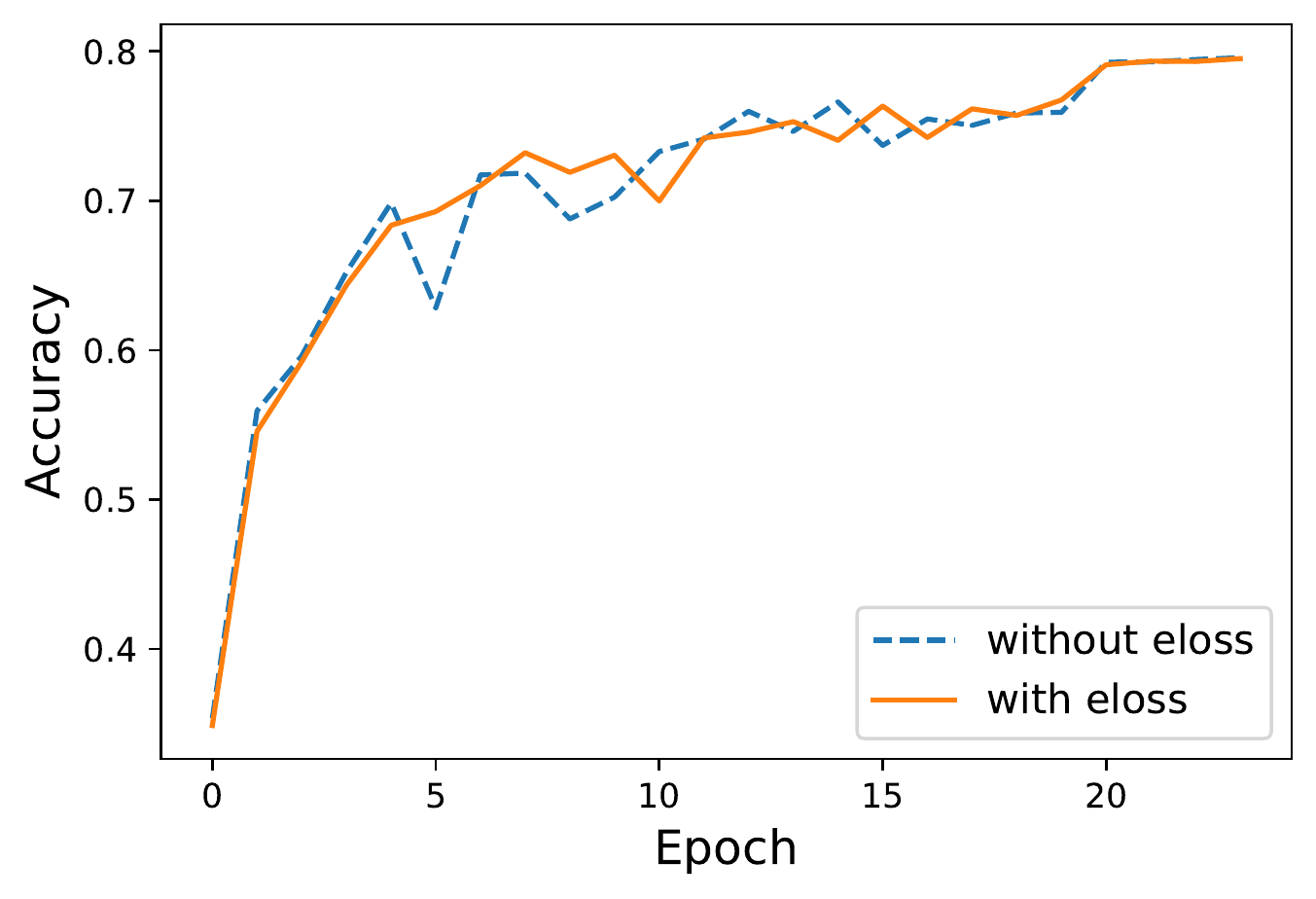}
          }
          \subfloat[mAP]{
            \includegraphics[width=0.333\textwidth]{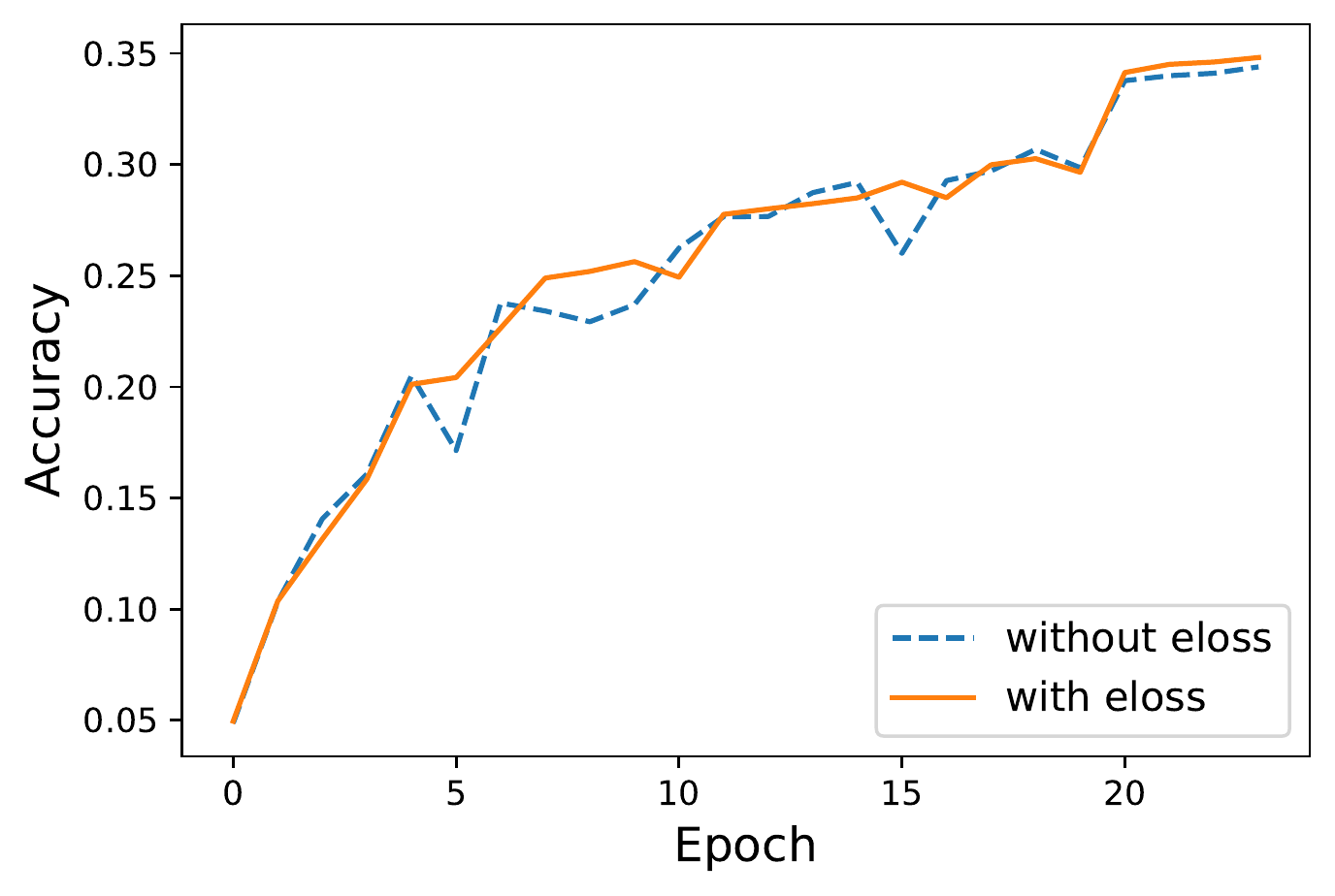}
          } 
          \subfloat[NDS]{
            \includegraphics[width=0.333\textwidth]{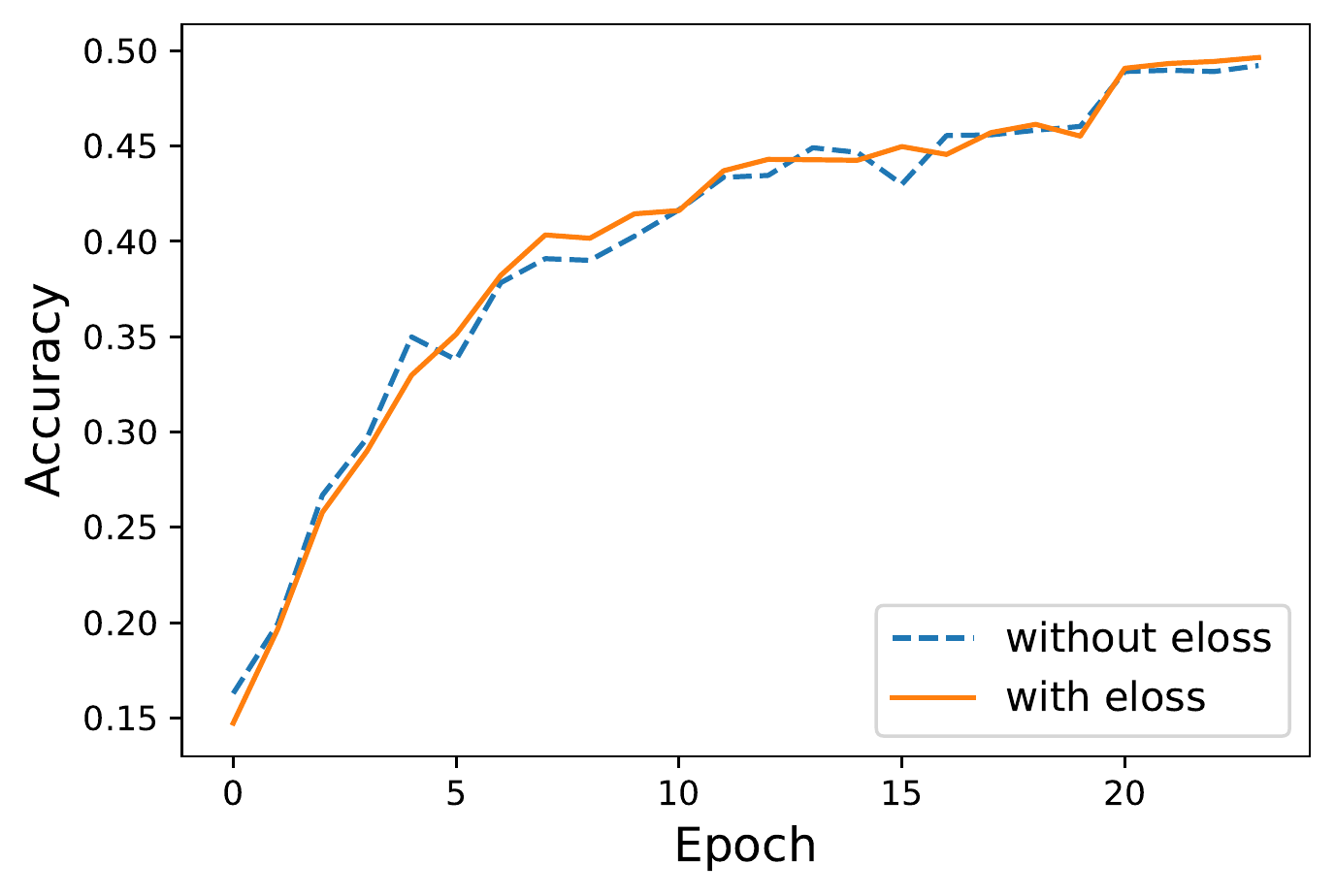}
          } 
          \caption{Convergences Curves of the model accuracy on NuSenes validation set for PointPillars  with or without Eloss. (a) the Average Precision of Car detection with Distance Threshold 1.0 meters; (b) mean Average Precision computed across 10 class of objects; (c) nuScenes detection score.}
          \label{training_process_comparison_figure}
        \end{figure*}

        To measure the impact of eloss on the model training process, we first conduct control experiments on the same model with and without eloss on the KITTI dataset and Nusenes dataset without noise. We plot the part of our experiment results in Figure~\ref{training_process_comparison_figure} to more intuitively show the impact of eloss on the volatility of the training process.
        
        To quantify the impact process, we use the mean absolute value slope (Mean Absolute Value Slope, MAVP) to measure the impact of eloss on the volatility of the model training curve and use the maximum precision index to measure the impact of eloss on the model training accuracy. The MAVP formula is as follows, where N is the number of sliding panes, and (k, k+1) refers to two adjacent time windows.

        \begin{equation}
            \mbox{MAVP}=\frac{1}{N}\sum_{k=1}^N(|x_{k+1}|-|x_k|))
        \end{equation}

        We applied the pointpillars and voxelnet models to the KITTI dataset to conduct the above control experiments. The experimental results are shown in the Table~\ref{kitti_eval_table}.

        \begin{table}[!ht]
            \centering
            \resizebox{0.97\columnwidth}{!}{
                \begin{tabular}{l|l|cc}
                    \hline
                    Model & Method & Max(\%) & MAVP(\%) \\ \hline
                    \multirow{3}*{PointPillars} & Without Eloss & 90.694 & 11.946 \\ 
                    ~ & With Eloss & 88.916 & 11.932 \\ 
                    ~ & Delta & -1.778 & \textbf{-0.014} \\ \hline
                    \multirow{3}*{VoxelNet} & Without Eloss & 94.873 & 10.959 \\  
                    ~ & With Eloss & 94.586 & 10.937 \\
                    ~ & Delta & -0.287 & \textbf{-0.022} \\ \hline
                \end{tabular}
            }
            \caption{The Car $AP_{R40}$ Max and MAVP of the models on KITTI validation set during the training process.}
            \label{kitti_eval_table}
        \end{table}

        The experimental results in the pivot table show that the maximum training accuracy decreases after adding eloss to both models. In terms of MAVP, the MAVP decreased after adding eloss, which means that the addition of eloss makes the above training process smoother.

        On the Nuscence dataset, we perform the above control experiments on the PointPillars model with three different metrics: Car $AP_dist1.0$, mAP, and NDS. The experimental results are shown in the Table~\ref{nus_eval_table}.

        \begin{table}[!ht]
            \centering
            \resizebox{0.97\columnwidth}{!}{
                \begin{tabular}{l|l|cc}
                    \hline
                    Metric & Method & Max(\%) & MAVP(\%) \\\hline
                    \multirow{3}*{Car $AP_{dist 1.0}$} & Without Eloss & 79.580 & 2.945 \\ 
                    ~ & With Eloss & 79.520 & 2.811 \\ 
                    ~ & Delta & -0.060 & \textbf{-0.135} \\ \hline
                    \multirow{3}*{mAP} & Without Eloss & 34.393 & 6.036 \\ 
                    ~ & With Eloss & 34.815 & 4.883 \\
                    ~ & Delta & \textbf{0.422} & \textbf{-1.153} \\ \hline
                    \multirow{3}*{NDS} & Without Eloss & 49.217 & 4.902 \\ 
                    ~ & With Eloss & 49.637 & 3.902 \\ 
                    ~ & Delta & \textbf{0.420} & \textbf{-1.000} \\ \hline
                \end{tabular}
            }
            \caption{The Max and MAVP of PointPillars on Nusenes validation set during the training process.}
            \label{nus_eval_table}
        \end{table}

        The table shows that the maximum accuracy of the Car category is lost during the training process after adding eloss. Still, the decline in MAVP shows that the addition of eloss moderates the volatility of the above training process. Similar observations for mAP, the average precision of multiple categories, and NDS, the Nusenes detection score, indicate that adding eloss to the model makes the training process smoother.

        The above is the experiment on the influence of eloss on model training without noise interference. In order to further understand the effect of eloss in the experiment, we will add Eloss to different parts of the network or conduct control experiments with anomalous data.

    \subsection{Varying Amount of Eloss}
        In this experiment, we apply continuous training to models that have already trained 80 epochs and see how adding Eloss to different fractions of the network may influence the model's accuracy on the KITTI test set.

        Eloss is a plug-and-play module, which means Eloss can be added to any place having a repetitive network structure and start functioning. For this reason, we can evaluate Eloss by considering different fractions of the network. In this experiment, we consider only repetitive structures grouped into blocks in SECOND backbone for convenience.
        
        After sorting out the experiment result of the best models, we get Table~\ref{eloss_amount_table}. 

        \begin{table}[!ht]
            \centering
            \resizebox{0.99\columnwidth}{!}{
                \begin{tabular}{l|l|l|ccc|c}
                    \hline
                    \multirow{2}*{Model} & \multirow{2}*{Epoch} & \multirow{2}*{Eloss} & \multicolumn{3}{c|}{Car $AP_{R40}$} &  \multirow{2}*{Time($ms$)}\\ 
                    ~ & ~ & ~ &Easy(\%) & Mod.(\%) & Hard(\%) \\ \hline \hline
                    \multirow{5}*{PointPillars} & 80 & 0 block & 84.20 & 74.82 & 68.02 & 4.50\\ 
                    ~ & 85 & 1 block & 82.65 & 71.30 & 64.66 & 7.41\\ 
                    ~ & 85 & 2 block & 80.90 & 69.79 & 64.34 & 16.74\\ 
                    ~ & 85 & 3 block & 80.23 & 68.92 & 63.49 & 34.88\\ 
                    ~ & best & 0 block & - & 77.60 & - & - \\ \hline \hline
                    \multirow{4}*{VoxelNet} & 80  & 0 block & 82.35 & 73.35 & 68.59 & 6.94\\ 
                    ~ & 85  & 1 block & 81.34 & 70.46 & 65.29 & 14.72\\ 
                    ~ & 85  & 2 block & 85.34 & 74.44 & 67.51 & 33.75\\ 
                    ~ & best & 0 blcok & - & 79.07 & - & -\\ \hline
                \end{tabular}
            }
            \caption{The accuracy for Car 3D detection of VoxelNet and PointPillars with different Eloss settings after continuous training on KITTI dataset.}
            \label{eloss_amount_table}
        \end{table}
        
        In these tables, there are gaps between the accuracy of models at 80 epoch and that of the fine-tuned models. Hence, a reasonable expectation is constructed that the model continually trains after 80 epochs. In the validation set, the accuracy raises after Eloss is applied as expected. However, in the test set, we get conflicting results.
        
        For PointPillars, in Table~\ref{eloss_amount_table}, the Car $AP_R40$ value is going down as the Eloss influences more SECOND blocks and is consistent with the previous experiment. The reason is that Eloss provides more constraints to the model parameters, making the model training process more difficult.
        
        In Table~\ref{eloss_amount_table}, the result for voxel-net differs from previous observations. When the Eloss influences one block of SECOND, the accuracy on the test set drops slightly, but when the Eloss influences two blocks of SECOND, the accuracy increases.
        
        For the abnormal observation of VoxelNet with Eloss, we undertake more experiments for further evaluation.

    \subsection{Comparison between Different Models}
        
        In this experiment, we use the same voxel encoder as VoxelNet, and blocks in SECOND where Eloss can influence are pre-trained with or without Eloss, and their parameters are frozen. Then we vary the size of the whole model, increasing the number of modalities and the degree of complexity. Result is shown in Table~\ref{comparision-table}.

        \begin{table*}[!ht]
          \centering
          \resizebox{0.99\textwidth}{!}{
              \begin{tabular}{l|l|ccc|ccc|ccc}
                \hline
                \multirow{2}*{Model} & \multirow{2}*{Method} & \multicolumn{3}{c|}{Car 3D $AP_{R40}$} & \multicolumn{3}{c|}{Cyclist 3D $AP_{R40}$} & \multicolumn{3}{c}{Pedestrian 3D $AP_{R40}$} \\
                &  &  Easy(\%)  & Mod.(\%) & Hard(\%) &  Easy(\%)  & Mod.(\%)  & Hard(\%) &  Easy(\%)  & Moderate(\%)  & Hard(\%) \\
                \hline
                SECOND\cite{yan2018second} & Without Eloss & 82.35 & 73.35  & 68.59 & 70.89 & 56.72 & 50.68 & 50.75 & 40.76 & 36.96\\
                 & With Eloss & 82.68 & 73.67  & 67.21 & 71.99 & 58.00 & 50.94 & 50.49 & 41.16 & 37.43 \\
                
                 & Delta & +\textbf{0.33} & +\textbf{0.32} & -1.38 & +\textbf{1.1} & +\textbf{1.28} & +\textbf{0.26} & -0.26 & +\textbf{0.4} & +\textbf{0.47} \\
                \hline
                +ResNet\cite{he2016deep} & Without Eloss & 80.29 & 67.37 & 60.94 & 75.70 & 52.37 & 46.10 & 39.64 & 31.13 & 28.95 \\
                 & With Eloss & 77.62 & 64.92 & 60.36 & 71.47 & 55.79 & 49.64 & 44.85 & 35.60 & 32.66 \\
    
                 & Delta & -2.67 & -2.45 & -0.58 & -4.23 & +\textbf{3.42} & +\textbf{3.54} & +\textbf{5.21} & +\textbf{4.47} & +\textbf{3.71}\\
                \hline
                +Correlation\cite{zheng2022multi} & Without Eloss & 73.47 &  62.47  & 57.99 & 63.08 & 49.55 & 44.33 & 42.46 & 35.11 & 32.16 \\
                +GNN\cite{scarselli2008graph} & With Eloss & 67.33 & 58.70  & 54.13 & 57.16 & 46.36 & 41.54 & 45.02 & 36.39 & 33.29  \\
    
                +FPN\cite{lin2017feature} & Delta & -6.14 & -3.77 & -3.86 & -5.92 & -3.19 & -2.79 & +\textbf{2.56} & +\textbf{1.28} & +\textbf{1.13} \\
                \hline
              \end{tabular}
            }
            \caption{Compare between the accuracy on test set for 3 different models with or without Eloss on 3 Class: Car, Cyclist, and Pedestrian, after 40 epochs train.}
          \label{comparision-table}
        \end{table*}

        In the SECOND+ResNet\cite{he2016deep} experiment, we take information from 2 modalities: point cloud and image. An increase in the accuracy of Cyclist and Pedestrian detection is observed. Both increase more than $3\%$, but there is a decrease in the accuracy of Car detection. 
        
        The situation is getting worse in the last model SECOND+RESNET+Correlation\cite{yan2018second,he2016deep,zheng2022multi}+GNN\cite{scarselli2008graph}+FPN\cite{lin2017feature}, only the accuracy of the Pedestrain detection increases. 
        
        These observations suggest that for a network with a voxel encoder, the more significant influence Eloss has, the better the network's performance on the object detection task.
        
        At this stage, we guess the reason for the above observation is that blocks influenced by Eloss are beneficial for compression as information is transmitted in those blocks, but the compression result is fragile and can be easily distorted by the latter inference process while training. Suggests that if Eloss is used, we need to consider using a lower learning rate for layers outside the Eloss coverage. We will take more effort into this topic in the future.



\section{Conclusion}

In this paper, we propose Eloss, an amplifier of the interpretability of a feature compression network based on the ideas behind the communication system. The Eloss is constructed by comparing the network-layer output with the information change expectations, which are generated according to the source coding. We implement Eloss to optimize a network in the direction converging to expectations. Through experiments in three different aspects, our study shows that training a 3D object detection model with Eloss is beneficial for both training speed and model interpretability. Having these results, we still see limitations in our work. For a model with a small fraction of network where Eloss can influence, adding Eloss can cause unexpected retention to the model-training process. In the future, we will investigate this limitation in detail and continue our research on the interpretability of models in intelligent driving.

\section*{Acknowledgements}
This work was supported by the National High Technology Research and Development Program of China under Grant No. 2018YFE0204300, and the National Natural Science Foundation of China under Grant No. U1964203,and sponsored by Meituan and Tsinghua University-Didi Joint Research Center for Future
Mobility.

\bibliography{reference.bib} 

\end{document}


\appendix

\section{Additional Implimentation Details}
We have discussed briefly in the main text about the implementation settings of our experiments, in this part we will give more information about our experiment settings, helping obtain results of our proposed methods.

\textbf{Code}. As described in the main text, our code will be made public under the Github repository link: \url{https://github.com/Discover304/Eloss}. We use PyTorch to construct our models, and using MMDetection3D\cite{mmdet3d2020} framework for data loading and report generating, 

\textbf{Hardware}. All our experiments undertaking on our internal server consisting of ten Nvidia RTX3090 GPUs and ten Nvidia RTX2080 GPUs. Due to our limited resources, all our models are not reaching the maximum performance as reported in the original work of those models. So, we control variables, keeping the number of training epochs the same to get the comparison experiment results as reasonable as we can. 

\textbf{Training parameters}. For KITTI\cite{geiger_2012_are, geiger2013vision} dataset, we use $50\%$-$50\%$ training-validation split. The batch size is set to 32. Use AdamW as the optimizer. All other setting is following the default config files of MVX-Net\cite{sindagi2019mvx} in MMDetection3D framework.

\begin{figure}[H]
    \centering
    \includegraphics[width=0.8\textwidth]{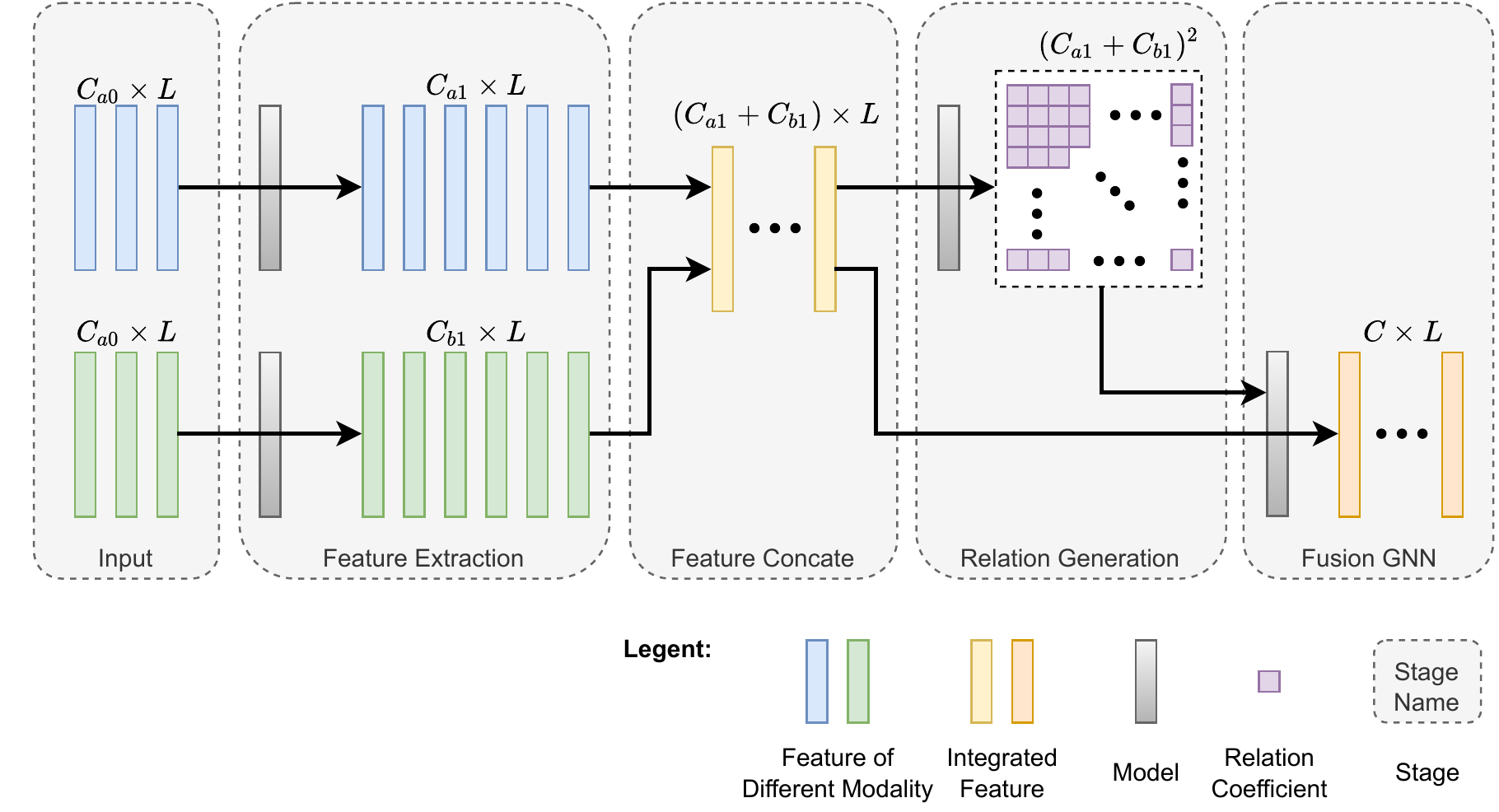}
    \caption{The whole structure of the model: SECOND+ResNet+Correlation+GNN+FPN.}
    \label{whole_structure}
\end{figure}

\section{Model Details}

There is a model we have mentioned but give little information about it. In this section we will give the detailed description of the model: SECOND\cite{yan2018second}+ResNet\cite{resnet2015kaiming}+Correlation\cite{zheng2022multi}+GNN\cite{scarselli2008graph}+FPN\cite{lin2017feature}. The whole model structure is as illustrated in Figure~\ref{whole_structure}.

\textbf{Feature extraction}. For a multi-modal fusion network, we need feature extraction network for all input modality, in this case, point cloud and image. So, we have the first two network structure, SECOND and ResNet. 

SECOND is the network which we have thoroughly discussed in the main text, for information compression of point cloud after voxel encoding. SECOND backbone network includes 3 repeated feature compression blocks, and each block includes 5 convolution layers. In addition, SECOND is where we modified to be influenced by Eloss. 

ResNet is for image information compression for later fusion tasks. We use ResNet-18 in real implementation. ResNet-50 is also applicable, and can get more plausible results, but it will consume more resources. Since we conduct only comparison experiments, we decide not use ResNet-50 as our image backbone, keeping it as an option for future work.

\textbf{Relation Generation}. We choose to use Graph Neural Network as a more sophisticated multi-modal fusion network compare with feature channel concatenation. So, to get an adjacency matrix to describe the feature level relation, we calculate the most basic type of relation, the linear correlation coefficient. We use this linear correlation coefficient matrix for later GNN fusion.

\textbf{Fusion GNN}. In this perception stage, there are two parts consist in fusion network, GNN and FPN. 

GNN or GCN to specific, considering a graph as a image, use convolution method in image similar way in graph scene. The core concept under GCN is learning a mapping function, to aggregate feature $f_i$ of node $v_i$ and its neighbor node features $x_j,. j\in N(v_i)$, generate the new representation of $v_i$\cite{kipf2017semi}. A single layer GCN can be described as following equation:

\begin{equation}
f_{n+1}=A\times (k*f_n+b)
\end{equation}

Where $f_n$ is the feature output of layer $n$, and $A$ is the adjacency matrix calculated in relation generation layer, $k$ and $b$ are trainable parameters.

FPN is abbreviation of feature pyramid network. In real application, GCN is enough for feature fusion, but as the number of layers of the GCN fusion network increases, the perceptive field also gradually increases, until each feature influenced by information about all relevant features, which leads to the convergence of the fused features. So, we use FPN to superimpose the outputs of different layers of the GCN, that is, the ensemble results with different perceptive field, as the final output of the GNN. 

\bibliographystyle{plain} 
\bibliography{reference} 

\section*{Checklist}

\begin{enumerate}

\item For all authors...
\begin{enumerate}
  \item Do the main claims made in the abstract and introduction accurately reflect the paper's contributions and scope?
    \answerYes{}
  \item Did you describe the limitations of your work?
    \answerYes{See Section 3.4 and Section 4.3.}
  \item Did you discuss any potential negative societal impacts of your work?
    \answerNo{}
  \item Have you read the ethics review guidelines and ensured that your paper conforms to them?
    \answerYes{}
\end{enumerate}

\item If you are including theoretical results...
\begin{enumerate}
  \item Did you state the full set of assumptions of all theoretical results?
    \answerYes{}
  \item Did you include complete proofs of all theoretical results?
    \answerYes{}
\end{enumerate}

\item If you ran experiments...
\begin{enumerate}
  \item Did you include the code, data, and instructions needed to reproduce the main experimental results (either in the supplemental material or as a URL)?
    \answerYes{See supplementary.}
  \item Did you specify all the training details (e.g., data splits, hyperparameters, how they were chosen)?
    \answerYes{See supplementary.}
  \item Did you report error bars (e.g., with respect to the random seed after running experiments multiple times)?
    \answerNo{Re-training for meaningful error bars is expensive.}
  \item Did you include the total amount of compute and the type of resources used (e.g., type of GPUs, internal cluster, or cloud provider)?
    \answerYes{See supplementary.}
\end{enumerate}

\item If you are using existing assets (e.g., code, data, models) or curating/releasing new assets...
\begin{enumerate}
  \item If your work uses existing assets, did you cite the creators?
    \answerYes{}
  \item Did you mention the license of the assets?
    \answerNo{Licenses are standard and can be found online.}
  \item Did you include any new assets either in the supplemental material or as a URL?
    \answerYes{Our code is available at  \url{https://github.com/Discover304/Eloss}}
  \item Did you discuss whether and how consent was obtained from people whose data you're using/curating?
    \answerNo{All datasets used in our work are publicly available.}
  \item Did you discuss whether the data you are using/curating contains personally identifiable information or offensive content?
    \answerNo{These discussions for KITTI dataset are available to public online.}
\end{enumerate}

\item If you used crowdsourcing or conducted research with human subjects...
\begin{enumerate}
  \item Did you include the full text of instructions given to participants and screenshots, if applicable?
    \answerNA{}
  \item Did you describe any potential participant risks, with links to Institutional Review Board (IRB) approvals, if applicable?
    \answerNA{}
  \item Did you include the estimated hourly wage paid to participants and the total amount spent on participant compensation?
    \answerNA{}
\end{enumerate}

\end{enumerate}